\documentclass{article}

\usepackage[final]{nips_2018}

\usepackage[utf8]{inputenc} %
\usepackage[T1]{fontenc}    %
\usepackage{hyperref}       %
\usepackage{url}            %
\usepackage{booktabs}       %
\usepackage{amsfonts}       %
\usepackage{nicefrac}       %
\usepackage{microtype}      %
\usepackage{hyperref}
\usepackage{url}
\usepackage{amsmath,amsthm}
\usepackage{graphicx}
\usepackage{xcolor}

\usepackage{titlesec}

\usepackage{amsfonts,bm}

\def\eqref#1{equation~\ref{#1}}

\def\1{\bm{1}}

\DeclareMathAlphabet{\mathsfit}{\encodingdefault}{\sfdefault}{m}{sl}
\SetMathAlphabet{\mathsfit}{bold}{\encodingdefault}{\sfdefault}{bx}{n}

\newcommand{\E}{\mathbb{E}}

\DeclareMathOperator*{\argmin}{arg\,min}

\title{BriarPatches: Pixel-Space Interventions \\
for Inducing Demographic Parity}

\author{
  Alexey A. Gritsenko,  Alex D'Amour,  James Atwood,  Yoni Halpern,  D. Sculley \\
  Google AI \\
  \texttt{(agritsenko|alexdamour|atwoodj|yhalpern|dsculley)@google.com} \\
}

\newcommand{\norm}[1]{\left\lVert#1\right\rVert}

\newcommand{\patch}{\mathrm{patch}}
\newcommand{\todo}[1]{}
\renewcommand{\todo}[1]{{\color{red} TODO: {#1}}}

\titlespacing*{\section}{0pt}{1.1\baselineskip}{\baselineskip}

\begin{document}

\maketitle

\vspace{-0.6cm}

\begin{abstract}
\vspace{-0.4cm}

We introduce the BriarPatch, a pixel-space intervention that obscures sensitive attributes from representations encoded in pre-trained classifiers. The patches encourage internal model representations not to encode sensitive information, which has the effect of pushing downstream predictors towards exhibiting demographic parity with respect to the sensitive information.  The net result is that these BriarPatches provide an intervention mechanism available at user level, and complements prior
research on fair representations that were previously only applicable by model developers and ML experts.
\end{abstract}

\begin{figure}[bh]
    \setlength{\tabcolsep}{0pt}
    \begin{tabular}{r r r r r }
    \includegraphics[width=0.23\textwidth, clip, trim=35 0 0 0]{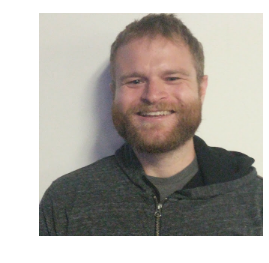} &
    \includegraphics[width=0.23\textwidth, clip, trim=35 0 0 0]{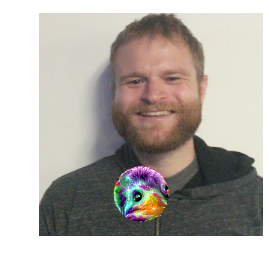} & 
    ~~~~~~~~~&
    \includegraphics[width=0.23\textwidth, clip, trim=35 0 0 0]{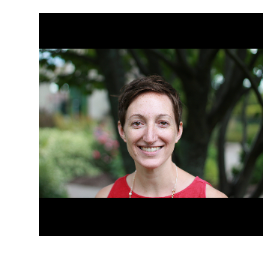} &
    \includegraphics[width=0.23\textwidth, clip, trim=35 0 0 0]{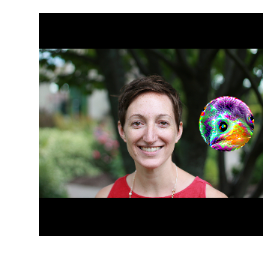} \\
    \scalebox{0.8}{$P(\texttt{man} | x) = 0.362$~~~~~} & 
    \scalebox{0.8}{$P(\texttt{man} | \patch(x)) = 0.002$~} & ~~~&
    \scalebox{0.8}{$P(\texttt{man} | x) = 0.115$~~~~~} & 
    \scalebox{0.8}{$P(\texttt{man} | \patch(x)) = 0.001$~} \\
    \scalebox{0.8}{$P(\texttt{woman} | x) = 0.005$~~~~~} & 
    \scalebox{0.8}{$P(\texttt{woman} | \patch(x)) = 0.005$~} & ~~~&
    \scalebox{0.8}{$P(\texttt{woman} | x) = 0.284$~~~~~} & 
    \scalebox{0.8}{$P(\texttt{woman} | \patch(x)) = 0.008$~} \\
    \end{tabular}
    \caption{Effect of a BriarPatch trained to remove perceived gender information.  Probabilities for the \texttt{man} and \texttt{woman} labels (see footnote 1 in Section~\ref{sec:results}) given by a pre-trained Inception V3 classifier are reduced after patch application. Images used with permission.}
    \label{fig:example_patch}
\end{figure}

\vspace{-0.6cm}
\section{Introduction: Fair Representation Learning from the Client Side}
\vspace{-0.5cm}

Recent papers in the machine learning fairness literature have proposed the idea of increasing
fairness by reducing or removing a deep network's ability to model a sensitive characteristic \citep{zemel2013learning,DBLP:journals/corr/abs-1802-06309,beutel2017data}.
These methods have been shown to be effective, but can only be employed by model developers at the
time that the model is trained.  From the perspective of an end user, some
additional mechanism may be useful on the client side to similarly reduce or remove the influence
of a sensitive characteristic in any downstream models that may make predictions based on that image.
This may be especially important for downstream models trained on data with strong
correlations between a sensitive characteristic, such as gender, and classification targets
such as {\tt athlete} the model is attempting to predict.

This paper presents a method that empowers end users to limit the influence of sensitive
characteristics, such as gender, in downstream image models. Building on recent work on adversarial instance generation \cite{brown2017adversarial}, the method takes the form of a {\em patch} of pixels that, when applied to an image, reduces the amount of information about a chosen sensitive characteristic contained in the representation computed using off-the-shelf models without unduly affecting the human perception of the image.

This approach can be viewed as inducing a fair representation of the image that ensures all downstream prediction tasks using the model output exhibit demographic parity.  We call this method for bias reduction a BriarPatch, and provide empirical results that suggest that it succeeds in reducing (but not removing) sensitive information from model representations, at the cost of reducing classifier utility.

\vspace{-0.3cm}
\paragraph{Related Work.}
\label{sec:related_work}
This paper builds of off prior work in the ML fairness literature around learning fair representations \cite{zemel2013learning}.
Previous work has shown empirically that fair representations ensure that downstream tasks satisfy certain fairness criteria regardless of the downstream actor's task \citep{DBLP:journals/corr/abs-1802-06309,beutel2017data}. Our approach extends this work by granting users, rather than algorithm designers, the agency to decide which (sensitive) attributes, if any, should be removed from model representations.

This paper also draws from prior work on adversarial instance generation and adversarial patches.
Adversarial examples are perturbations applied to inputs in order to change classifier predictions \cite{intriguing_properties}. These may be imperceptible to humans \cite{explaining_harnessing} or
may take the form of a small visible alteration (e.g. a circular array of pixels) alters classifier output when pasted onto an image \cite{brown2017adversarial}.

\vspace{-0.4cm}
\section{Setup and Problem Formulation}
\label{sec:setup}
\vspace{-0.5cm}

We consider a multi-label image classification problem where the task is to identify the presence or absence of $k$ concepts in each image.
We consider two distinct parties involved in classifying images, defined first in \cite{dwork2012fairness}, and later extended in work on fair representation learning \citep{zemel2013learning,DBLP:journals/corr/abs-1802-06309,beutel2017data}.
The first party is the \emph{data owner}, a trusted party that holds user data and uses it to construct a pre-trained encoder that yields vector representations of images. The second party is the \emph{vendor}, who uses these representations in some downstream task, such as predicting the label
{\tt athlete} for the given image.

Concretely, let $X$ be a random image drawn from a given population of images taking values $x$ in support $\mathcal X$.
The classification task is to map an image $x$ to a binary $k$-vector of concept labels $\hat y(x) \in \{0, 1\}^{k}$.
The data owner constructs an encoder that maps an image $x$ to a $k$-vector of logits $r(x)$, such that each element of the vector, $r_\ell(x)$ for $\ell = 1, \cdots k$, corresponds to an estimate of $\mathrm{logit} P(Y_l = 1 \mid x)$, where $\mathrm{logit}(p) = \log(\frac{p}{1-p})$.  The representation $r(x)$ is then passed on to the vendor.  For a given image $x$, the vendor uses this representation $r(x)$ to make decisions as they please. For example, in our experiments, $r(x)$ is the logits layer of an Inception V3 network trained on OpenImages V1.

We assume that the end user who provided each image has some sensitive attribute that they intend to obscure from the vendor.
Let $A$ be the sensitive attribute of the end user who provided a randomly drawn image, taking values $a$ in support $\mathcal A$. 
We will assume that $A$ is binary, so that $\mathcal A = \{0, 1\}$.
Additionally, we will assume that $A$ appears in the label set, at index $\ell(A)$, so that $r_{\ell(A)}(x)$ corresponds to an estimate of $\mathrm{logit} P(Y_{l(A)} = 1 \mid x)$.

We consider the problem of designing a \emph{patch} that end users can apply to their images in order to obscure their sensitive information from the prediction vendor.  Specifically, we design a transformation $\patch(x)$ to induce certain statistical parities in the representation $r(\patch(X))$ when the patch is applied across the population.

\vspace{-0.4cm}
\section{Patch Models}
\vspace{-0.5cm}

Our goal is to design a patch transformation so that the sensitive attribute $A$ cannot be recovered from the representation $r(\patch(x))$.
In a similar approach to \cite{DBLP:journals/corr/abs-1802-06309}, we frame the problem in terms of thwarting an adversarial vendor who attempts to recover $A$ from $r(\patch(X))$ using some classifier $h$ while maintaining the utility of the original multi-class classifier $f$. 

\vspace{-0.3cm}
\paragraph{General case} In full generality, we seek a patch transformation, $\patch(\cdot)$ corresponding to the following objective:
\begin{equation}
\label{eq:objective}
\argmin_{\patch \in \mathcal P} \max_{h \in \mathcal H} \E[\mathcal L_A\{A, h\circ r(\patch(X))\}
+ \lambda \mathcal L_R\{Y_{-\ell(A)}, f \circ r(\patch(X))\}].
\end{equation}
Here, $\mathcal L_A$ and $\mathcal L_R$ are classification losses for predicting $A$ and $Y$ respectively.
The $\mathcal L_A$ term corresponds to the adversarial vendor's objective, whereas the $\mathcal L_R$ term is a regularizer that constrains the patch to trade off thwarting the adversary with maintaining the utility of the original classifier, $f \circ r(X)$.

Here, we define the family of patch transformations $\mathcal P$ as a set of image transformations parameterized by a patch -- a small circular array of pixels.  Following \cite{brown2017adversarial}, the patch transformation, $\patch(\cdot)$, randomly rotates the patch and overpaints it on a random location on the image. 
Thus, a patch corresponding to the $\patch(\cdot)$ transformation that successfully solves the optimization problem should have properties that are, in expectation, translation and rotation invariant.

\vspace{-0.3cm}
\paragraph{Linear adversary and relaxations}
In this work, we consider a special case of Eq.~\ref{eq:objective} where the sensitive attribute $A$ is binary, and the family of adversarial classifiers $\mathcal H$ is constrained to be linear. %

In addition, we relax the objective in Eq.~\ref{eq:objective} in two ways.
First, we replace the adversarial optimization of $\mathcal L_A$ with a maximum mean discrepancy (MMD) \cite{gretton2012kernel}:
$$\max_{\mathcal H} \left[E[h\circ r(\patch(X)) \mid A = 1] - E[h \circ r(\patch(X)) \mid A = 0])\right].$$ 
When $\mathcal H$ is the set of linear functions, this MMD has a closed form, and reduces to the $L_2$ distance between the group means.
Secondly, we replace the original classifier loss $\mathcal L_R$ with the $L_2$ distance between the un-patched and patched image representations. This regularizer encourages the patched logits of non-sensitive attributes to remain close to their un-patched values, with the goal of preserving some of the utility of the original classifier $f$. This yields the following objective:
\begin{equation}
    \argmin_{\patch \in \mathcal P} \norm{E[r(\patch(x))|A=1] - E[r(\patch(x))|A=0]}_2^2+\lambda\E[\norm{r(x) - r(\patch(x))}^2_2].
    \label{eq:linear adversary}
\end{equation}
Figure \ref{fig:example_patch} shows an example patch trained according to this objective.

\vspace{-0.4cm}
\section{Experimental Results and Discussion}
\label{sec:results}
\vspace{-0.5cm}

\begin{figure}
    \centering
    \begin{tabular}{c c c}
        \includegraphics[width=0.33\textwidth,]{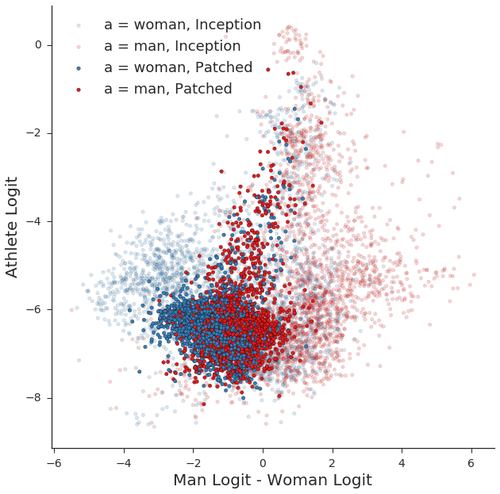} & 
        \qquad&
        \includegraphics[width=0.38\textwidth,]{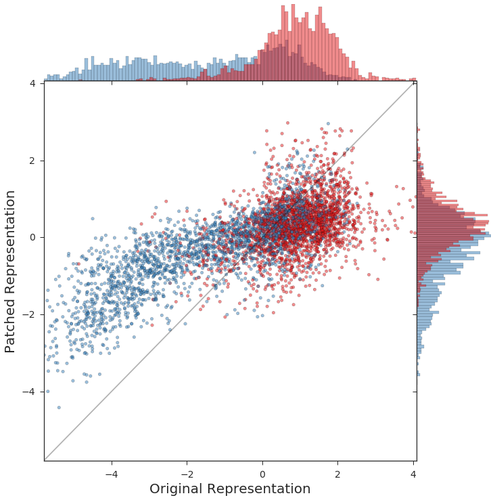} \\
        (\textbf{a}) Logit Representation & \qquad & (\textbf{b}) Adversarial Logits
    \end{tabular}
    \caption{Effects of patches on representations. (\textbf{a}) Image representation change in the ({\tt man}, {\tt woman}, {\tt athlete}) logit space. (\textbf{b}) $\mathrm{logit}\hat P(A=\texttt{man} \mid r(x))$ from an adversarial linear classifier before (x-axis) and after (y-axis) applying the patch.}
    \label{fig:patch 30 reg}
    \vspace{-0.4cm}
\end{figure}

In our experiments we sought to remove perceived gender information from the predictions made by an Inception V3 model trained on the OpenImages V1 dataset. This model predicts 6012 binary labels for a given image, including labels for {\tt man} and {\tt woman},\footnote{Although these provided labels are used as placeholders for perceived gender for these experiments, we acknowledge the limitations of this approach and recognize that gender is inherently a non-binary concept.} a number of labels that strongly correlate with it in the dataset (e.g. {\tt athlete}).

To explore the simplest case, we limited ourselves to de-biasing representations formed by concatenating the logit of a single binary target label (e.g. {\tt athlete}) and the gender logits {\tt man} and {\tt woman}. We measured success in removing gender information by AUC-ROC of a binary gender logistic regression classifier trained on representations formed by patched images.  We measured classifier utility on the single binary target label with the Average Precision (AP) metric. Patches were trained and evaluated on disjoint gender-balanced subsets of images from the OpenImages V3 validation set, which contained either the {\tt woman} or the {\tt man} label, but not both simultaneously. Patches were trained for $200$ epochs using
SGD with a learning rate of $10^{-3}$ and a batch size of $128$.

We measured the effect of the patch with several metrics related to classifier performance and fairness, and also explored the qualitative effect of patch application on downstream representations.

\vspace{-0.25cm}
\paragraph{Representation effects}
In Figure~\ref{fig:patch 30 reg}, we show the representation-space transformation $r(x) \mapsto r(\patch(x))$ induced by a patch.
To remove distinctions between images with {\tt man} and {\tt woman} labels, the patch maps representations toward the boundary between the regions primarily occupied by {\tt man} and {\tt woman} images in the original representation.
We also show the effects of the patch on the logits of an adversarial linear classifier attempting to recover the gender attribute from the representation.
The patch makes these logits less separable between the {\tt man} and {\tt woman} classes, although it does not make their distributions indistinguishable.
Importantly, the logits are squeezed toward zero for most images, reducing the certainty with which an adversary can classify individual images on the basis of the image representation.

\begin{figure}
    \centering
    \setlength{\tabcolsep}{0pt}
    \begin{tabular}{c c c c c}
        \includegraphics[width=0.48\textwidth]{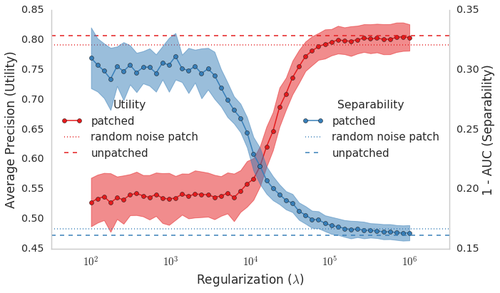} &
        ~~~ &
        \includegraphics[width=0.24\textwidth]{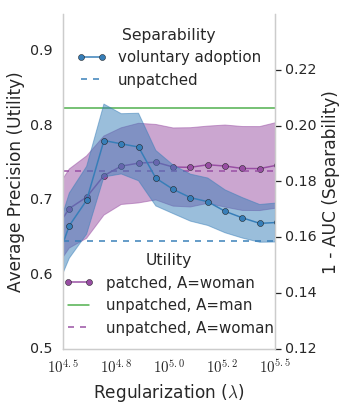} &
        ~~~ &
        \includegraphics[width=0.24\textwidth]{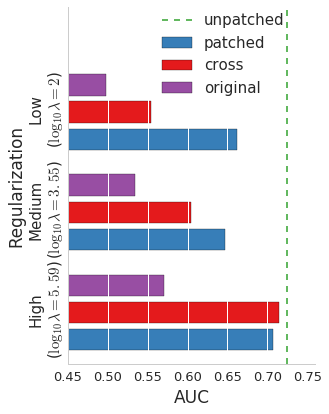} \\
    \end{tabular}
    \caption{The effect of BriarPatch on classifier output. Results shown for the {\tt athlete} label, but are representative of other labels. \textbf{Left:} Utility-separability trade-off for patches trained with different levels of regularization. Metric standard deviations (shown as shaded areas) were obtained by training $10$ different patches per regularization and evaluating each of them $25$ times on each image. Results for random noise patch are provided as a baseline for a patch that randomly masks out a part of the image. \textbf{Center:} When applying the patch only to images with the {\tt woman} label, we find a regime around $\log_{10}\lambda=4.69$ where separability is decreased but utility is unaffected.  \textbf{Right:} Gender prediction accuracy using the original Inception classifier ($\mathrm{logit}(\tt{man}) - \mathrm{logit}(\tt{woman})$; original), classifiers trained on the unpatched and patched representations (unpatched and patched respectively), and a classifier trained on the unpatched representation used to make prediction on the patched images (cross).}
    \label{fig:tradeoff}
    \vspace{-0.6cm}
\end{figure}

\vspace{-0.25cm}
\paragraph{Separability and Utility}
Varying the regularization amount $\lambda$ allowed us to assess the separability-utility trade-off attainable by the patch. Figure~\ref{fig:tradeoff}~(left) shows that the patch undergoes a regime switch, where it transitions from having little effect on the classifier's ability to predict the {\tt athlete} label on the test set (right side of the plot) and little success in de-biasing the logit representation to reducing the amount of the sensitive gender information at the cost of reducing the classifier utility (left side of the plot; $\Delta\mathrm{AUC}\approx0.16$ and $\Delta\mathrm{AP}\approx-0.27$.

Interestingly, as shown in Figure~\ref{fig:tradeoff}~(center), there is a regime where separability is decreased but utility remains unaffected for images with the {\tt woman} label. This suggests that voluntary application of the patch can improve fairness with minimal utility loss.
\footnote{
The patch as it is currently trained is not optimized for this use case.
When patch application is voluntary, increasing the patch strength too much (low regularization) leads to conspicuous representation changes that allow the vendor to detect patch application; meanwhile, decreasing the patch strength too much (high regularization) does not allow the patch to obscure sensitive characteristics effectively (see Figure~\ref{fig:representation-woman-only}). We speculate that this accounts for the peak in Figure~\ref{fig:tradeoff}~(center).
Patches tailored to voluntary application could potentially eliminate this tension.}

\vspace{-0.25cm}
\paragraph{Effects on Demographic Parity} BrairPatch controls the worst-case downstream demographic disparity by minimizing an adversarial vendor's ability to recover the sensitive attribute from patched representations \cite{DBLP:journals/corr/abs-1802-06309}. This guarantees that any other vendor (e.g. a vendor predicting the {\tt athlete} label) will have a smaller demographic disparity than the adversarial one. When the patch is unable to remove \textit{all} sensitive information from $r(x)$, this guarantee permits the patch to reduce the vendor's demographic parity as shown in Figure~\ref{fig:demographic-discrepancy-bound}.

\vspace{-0.25cm}
\paragraph{Confusing the original gender classifier}
We found that even when trained with little or no regularization, the patch intervention was unable to completely remove gender information from the considered representations (see Figure~\ref{fig:tradeoff}). This means that a determined vendor may use these representations to recover the gender from the obfuscated logits. However, we also found that the trained patches succeeded in confusing a na\"{i}ve vendor that directly used the original {\tt man} and {\tt woman} logits in the patched representation for recovering the gender ($\mathrm{AUC}\approx0.5$; Figure~\ref{fig:tradeoff}~(right)).

\vspace{-0.25cm}
\paragraph{Acknowledgements} We would like to thank Eric Breck, Erica Greene, and Shreya Shankar for contributing to earlier versions of this project and Pallavi Baljeka for insightful discussions.

\bibliographystyle{unsrt}
\bibliography{main}

\vfill

\pagebreak

\appendix

\section{Appendix}

\begin{figure}[ht]
    \centering
    \setlength{\tabcolsep}{0pt}
    \begin{tabular}{c c c c c}
        \includegraphics[width=0.30\textwidth]{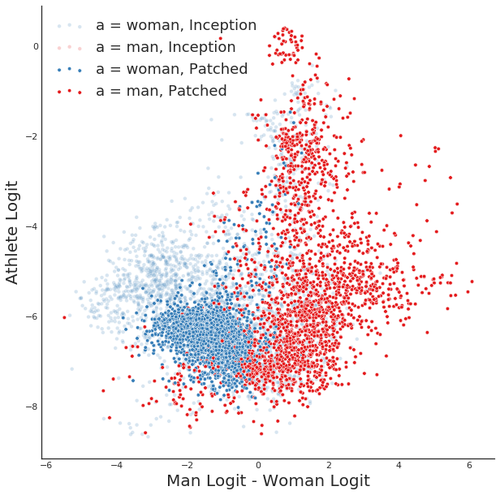} &
        ~~~ &
        \includegraphics[width=0.30\textwidth]{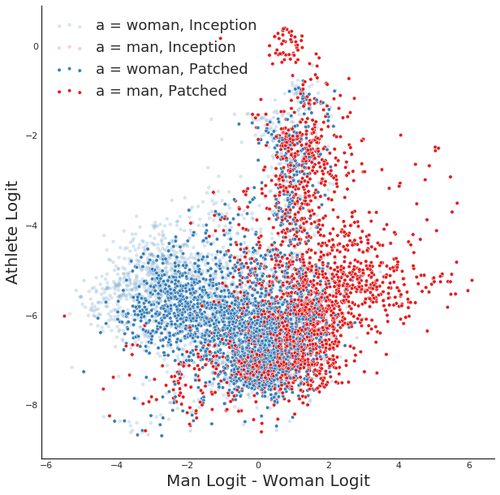} &
        ~~~ &
        \includegraphics[width=0.30\textwidth]{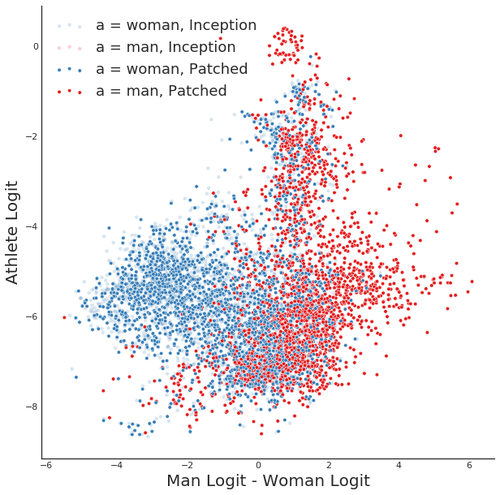} \\
    \end{tabular}
    \caption{Evolution of logit representations for patches with regularisation levels corresponding to points left of- (left; $\log_{10}\lambda=3.55$), on the- (center; $\log_{10}\lambda=4.69$) and right of (right; $\log_{10}\lambda=5.59$) the separability peak in Figure~\ref{fig:tradeoff}~(center).}
    \label{fig:representation-woman-only}
\end{figure}

\begin{figure}[ht]
    \centering
    \setlength{\tabcolsep}{0pt}
    \begin{tabular}{c c c}
        \includegraphics[width=0.60\textwidth]{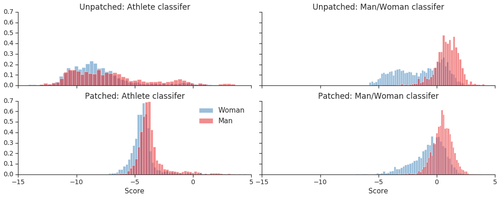} &
        ~~~ &
        \includegraphics[width=0.36\textwidth]{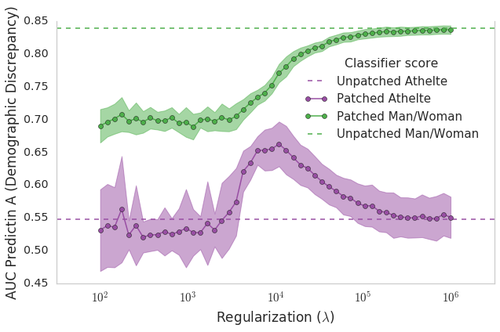} \\
        (\textbf{a}) (Adversarial) vendor classifier score distributions. & \qquad & (\textbf{b}) Demographic discrepancy bound.
    \end{tabular}
    \caption{Effects on downstream demographic parity. \textbf{(a)} Distribution of scores for a classifier trained to predict the sensitive gender attribute (i.e. an adversarial vendor; right column) and a classifier trained to predict the {\tt athlete} label ({\tt athlete} vendor; left column) on the original (top row) and patched (bottom row) representations shown for a single patch. Differences between the score distributions reflect the demographic parity gap between the sensitive groups. Applying the patch makes the gender classifier scores less separable between the groups, but does not necessarily have the same effect on the {\tt athlete} classifier scores. \textbf{(b)} AUC for predicting the sensitive gender attribute using the adversarial vendor scores (green), and using the {\tt athlete} scores (purple); unpatched baselines are shown as dashed lines. The adversarial vendor's AUC is consistently above that of the {\tt athlete} vendor, and serves as an upper bound that BriarPatch minimizes. However, subject to this upper bound, the demographic discrepancy of the vendor may increase (e.g. around $\log\lambda=4$). At low regularization the vendor's demographic discrepancy is decreased beyond the baseline and approaches the optimal value of $\mathrm{AUC}=0.5$.}
    \label{fig:demographic-discrepancy-bound}
\end{figure}

\end{document}